\documentclass[conference,letterpaper]{IEEEtran}
\usepackage[letterpaper, left=1in, right=1in, bottom=1in, top=0.75in]{geometry}

\usepackage{CJKutf8} 

\usepackage{graphicx,xcolor}  
\usepackage{url}
\usepackage{multirow}

\title{ChatGPT as a Translation Engine: A Case Study on Japanese-English}
\author{%
  Vincent Michael Sutanto${}^{1}$　Giovanni Gatti De Giacomo${}^{1}$　Toshiaki Nakazawa${}^{2}$　Masaru Yamada${}^{3}$\\
${}^{1}$Yaraku, Inc.　${}^{2}$The University of Tokyo　${}^{3}$Rikkyo University\\ \texttt{\{vincent, giovanni\}@yaraku.com, nakazawa@nlab.ci.i.u-tokyo.ac.jp, }\\
\texttt{masaru.yamada@rikkyo.ac.jp}
}

\begin{document}
\begin{CJK}{UTF8}{min} 

\maketitle

\begin{abstract}
This study investigates ChatGPT for Japanese-English translation, exploring simple and enhanced prompts and comparing against commercially available translation engines. 
Performing both automatic and MQM-based human evaluations, we found that document-level translation outperforms sentence-level translation for ChatGPT. 
On the other hand, we were not able to determine if enhanced prompts performed better than simple prompts in our experiments. 
We also discovered that ChatGPT-3.5 was preferred by automatic evaluation, but a tradeoff exists between accuracy (ChatGPT-3.5) and fluency (ChatGPT-4).
Lastly, ChatGPT yields competitive results against two widely-known translation systems.
\end{abstract}

\section{Introduction}
Recently, ChatGPT has emerged as a versatile tool, finding applications in several domains due to its multi-functional capabilities. 
Beyond its extensive utility, ChatGPT extends its prowess to include translation tasks, showcasing its adaptability in bridging language barriers \cite{kocmi-etal-2023-findings}.

Interestingly, ChatGPT, as a machine translation tool, offers more than just standard translation; it can be further customized by "enhancing" the prompt, producing results that are preferred by professional translators \cite{yamada-optimizing-2023}.
This feature empowers users to curate translations that align more precisely with their intended domain, purpose and tone \cite{yamada-optimizing-2023} \cite{peng-etal-2023-towards}.

However, a comprehensive investigation into the use of ChatGPT as an MT system is still required to unveil its applicability. Several questions remain unclear: 
does a general-purpose model like ChatGPT perform as well as specialized translation engines trained specifically for translation tasks? 
Is translating the entire document at once better than translating sentence by sentence?
How do simple and enhanced prompting techniques affect the translation results? 
Is there any discernible difference between ChatGPT-3.5 and ChatGPT-4? 
Answering these types of questions will help us understand and provide valuable insights into ChatGPT's role in translation technology.

In this paper, we employed ChatGPT-3.5 and ChatGPT-4 APIs \footnote{\textit{gpt-3.5-turbo} and \textit{gpt-4} models} to answer the previously mentioned research questions. Additionally, we evaluate the performance against the two most widely used commercial systems for JA-EN translation.
Our investigation reveals the following:

\begin{enumerate}
\item ChatGPT makes better translations when it handles whole documents instead of sentence by sentence, likely due to better context awareness as a translation system.
\item We could not find conclusive evidence with our experiments that enhanced prompts lead to higher quality translations. 
However, this could be due to the small-scale of our human evaluation and the bias towards gold data in the automated evaluations. 
Therefore, we believe larger scale human evaluations are required to accurately assess the quality difference.
\item ChatGPT-3.5 emerged as the favored model compared to ChatGPT-4 in automatic evaluation, but human evaluation reveals that the end user can choose between accuracy (ChatGPT-3.5) or fluency (ChatGPT-4).
\item For the JA-EN pair translation, both ChatGPT 3.5 and 4 performed competitively against two commercial MT systems, solidifying their position as viable translation system options.
\end{enumerate}

\section{Methodology}
\subsection{Datasets}
We used several publicly available JA-EN datasets for our experiments: ParaNatCom \cite{paranatcom_utiyama}, FLORES \cite{flores-nllb2022}, Novels \cite{utiyama2003english}, KFTT \cite{neubig11kftt}, and WMT News \cite{tiedemann-2012-parallel}. 
These datasets were handpicked to investigate translation quality on a variety of domains, lengths, and styles. 
Due to budget constraints for the API, we sampled five documents from each dataset for the automatic evaluation.
Additional details about the datasets are provided in Appendix \ref{appendix:dataset}.

\subsection{Prompting}

\begin{table}[t]
\centering
\caption{Simple Prompt for ChatGPT}
\small 
\label{tab:method-prompt-simple}
\begin{tabular}{llcc}
\hline
\textit{Translate this document from \textless src\_language\textgreater{} to \textless tgt\_} \\
\textit{language\textgreater{}:} \\
\textit{\textless text\textgreater{}}\\
\hline
\end{tabular}
\end{table}

\begin{table}[t]
\centering
\caption{Enhanced Prompt for ChatGPT}
\small 
\label{tab:method-prompt-enhanced}
\begin{tabular}{llcc}
\hline
\textit{You are a translation engine. Perform the following}\\
\textit{steps carefully:} \\
\textit{1. - Translate the following  \textless src\_language\textgreater{} document}\\
\hspace{0.6cm}\textit{delimited by triple backticks to \textless tgt\_language\textgreater{}}\\
\textit{2. - The translation is expected to be in the field of}\\
\hspace{0.6cm}\textit{\textless field\textgreater{}.} \\
\textit{3. - The expected output is expected to have \textless style\textgreater{}}\\
\hspace{0.6cm}\textit{style.} \\
\textit{4. - Change the translation tone into  \textless tone\textgreater{}.} \\
\textit{```} \\
\textit{\textless text\textgreater{}}\\
\textit{```} \\
\textit{After finishing those steps, return the final result only!}\\

\hline
\end{tabular}
\end{table}

In this work, we investigated two prompting strategies: \textit{Simple} and \textit{Enhanced} prompts.
The \textit{Simple} prompt simply directs ChatGPT to translate text based on source and target languages (Table \ref{tab:method-prompt-simple}). 
On the other hand, the \textit{Enhanced} prompt, motivated by \cite{yamada-optimizing-2023}, instructs ChatGPT to consider category, style, and tone as well as the source and target languages for a more customized output (Table \ref{tab:method-prompt-enhanced}).

With the help of professional translators, we assigned distinct categories, styles, and tones to each dataset, ensuring the translation output aligns with the specific characteristics of each dataset (Appendix \ref{appendix:enhanced-prompt}). 

\subsection{Automatic Evaluation}
We computed three commonly used automatic evaluation metrics to measure overall translation quality: BLEU \cite{papineni-etal-2002-bleu}, COMET \cite{rei-etal-2020-comet}, and DA-BERT \cite{zhan-etal-2021-difficulty}.

\subsection{Human Evaluation}
We also conducted a human evaluation to appraise machine translation quality using the Multidimensional Quality Metrics (MQM) framework\footnote{\url{https://themqm.org/}}. 
The evaluation was conducted by two professional translators, both with background and experience in JA-EN translation. 
For the human evaluation, one sample from the document-level translation of each of the following datasets were used: ParaNatCom \cite{paranatcom_utiyama}, Novels \cite{utiyama2003english}, and WMT News \cite{tiedemann-2012-parallel}. 

To facilitate the annotation process, we designed an evaluation tool for the MQM framework that we are open sourcing\footnote{\url{https://github.com/yaraku/he-tool}}. 
We hope that the tool will aid researchers and practitioners in conducting similar human evaluations for their machine translation studies. 
Finally, the tool also comes pre-configured with categories and weights selected by linguists for evaluating both Japanese and English sentences.

\vspace{0.2cm}
\section{Results and Discussion}

\begin{table*}[ht]
\centering
\caption{BLEU, COMET, and DA-BERT results per dataset (higher is better).}
\small 
\label{tab:rnd-results}
\begin{tabular}{llcccccc}
\hline
\multirow{2}{*}{Dataset} & \multirow{2}{*}{System} & \multicolumn{2}{c}{BLEU} & \multicolumn{2}{c}{COMET} & \multicolumn{2}{c}{DA-BERT}\\
&& Doc & Sent & Doc & Sent & Doc & Sent\\
\hline
\multirow{4}{*}{ParaNatCom} & ChatGPT-3.5 Simple	&36.39 &	36.07 &	0.906 &	0.903 &	\textbf{0.190} &	0.193 \\
& ChatGPT-4 Simple	    &37.55 &	38.09 &	0.907 &	0.903 &	0.181 &	0.185 \\
& ChatGPT-3.5 Enhanced	&37.64 &	34.11 &	0.907 &	0.905 &	0.178 &	\textbf{0.200} \\
& ChatGPT-4 Enhanced	&38.25 &	36.45 &	0.901 &	0.891 &	0.184 &	0.199 \\
\cline{2-8}
& Commercial System A      &\textbf{49.14} &	\textbf{49.32} &	\textbf{0.911} &	\textbf{0.912} &	0.172 &	0.156 \\
& Commercial System B      &47.60 &	47.76 &	\textbf{0.911} &	0.911 &	0.148 &	0.148 \\
\hline
\multirow{4}{*}{FLORES} & ChatGPT-3.5 Simple	&31.82 &	33.70 &	0.903 &	0.884 &	0.198 &	0.180 \\
& ChatGPT-4 Simple	    &31.82 &	\textbf{35.43} & \textbf{0.904} &	0.870 &	0.187 &	0.188 \\
& ChatGPT-3.5 Enhanced	&34.07 &	27.84 &	0.899 &	0.858 &	0.187 &	0.211 \\
& ChatGPT-4 Enhanced	&29.28 &	27.52 &	0.891 &	0.856 &	0.194 &	0.214 \\
\cline{2-8}
& Commercial System A                 &30.19 &	32.62 &	0.895 &	\textbf{0.899} &	\textbf{0.221} &	0.199 \\
& Commercial System B      &\textbf{41.43} &	32.94 &	0.893 &	0.773 &	0.176 &	\textbf{0.217} \\
\hline
\multirow{4}{*}{KFTT} & ChatGPT-3.5 Simple	&18.52 &	16.69 &	0.846 &	0.835 &	0.031 &	0.026 \\
& ChatGPT-4 Simple	    &17.52 &	18.51 &	0.843 &	0.836 &	0.018 &	0.024 \\
& ChatGPT-3.5 Enhanced	&17.79 &	14.60 &	\textbf{0.848} &	0.837 &	0.024 &	\textbf{0.062} \\
& ChatGPT-4 Enhanced    &16.62 &	15.23 &	0.845 &	0.830 &	0.018 &	-0.002\\
\cline{2-8}
&Commercial System A                  &19.51 &	19.36 &	0.842 &	0.840 &	0.027 &	\textbf{0.062} \\
&Commercial System B       &\textbf{20.32} &	\textbf{20.23} &	0.844 &	\textbf{0.843} &	\textbf{0.062} &	0.058 \\
\hline
\multirow{4}{*}{Novels} & ChatGPT-3.5 Simple	&12.79 &	13.26 &	0.860 &	0.840 &	0.272 &	0.274 \\
& ChatGPT-4 Simple	    &14.65 &	14.19 &	0.857 &	0.837 &	0.271 &	0.272 \\
& ChatGPT-3.5 Enhanced	&13.69 &	10.48 &	\textbf{0.865} &	0.787 &	0.274 &	\textbf{0.297} \\
& ChatGPT-4 Enhanced    &14.90 &	12.07 &	0.829 &	0.799 &	0.272 &	0.279 \\
\cline{2-8}
& Commercial System A                  &\textbf{15.49} &	12.97 &	0.864 &	\textbf{0.855} &	\textbf{0.280} &	0.293 \\
& Commercial System B       &14.64 & \textbf{14.85} &	0.847 &	0.845 &	0.267 &	0.268 \\
\hline
\multirow{4}{*}{WMT News} & ChatGPT-3.5 Simple	&25.84 &	26.69 &	0.819 &	0.819 &	0.185 & \textbf{0.199} \\
& ChatGPT-4 Simple	    &26.87 &	28.05 &	0.820 &	0.819 &	0.187 &	0.191 \\
& ChatGPT-3.5 Enhanced	&24.06 &	25.30 &	0.822 &	0.816 &	0.186 &	0.198 \\
& ChatGPT-4 Enhanced    &17.55 &	16.96 &	0.800 &	0.787 &	0.139 &	0.176 \\
\cline{2-8}
& Commercial System A                 &16.75 &	24.53 &	0.702 &	0.805 &	-0.047 &	0.174 \\
& Commercial System B      &\textbf{41.02} &	\textbf{41.55} &	\textbf{0.832} &	\textbf{0.832} &	\textbf{0.189}  &	0.188 \\
\hline
\end{tabular}
\end{table*}

\begin{table}[ht]
\centering
\caption{MQM scores for each system (lower is better).}
\small
\label{tab:rnd-human-results}
\begin{tabular}{lccc}
\hline
\multicolumn{1}{c}{System} & MQM & Accuracy & Fluency   \\
\hline
ChatGPT-3.5 Simple      & 84.50          & \textbf{55.67} &	26.00 \\
ChatGPT-4 Simple	    & 95.83          & 82.67	      & 11.83 \\
ChatGPT-3.5 Enhanced	& \textbf{83.17} & 59.67          & 17.00 \\
ChatGPT-4 Enhanced	    & 103.25         & 81.00          &	15.50 \\
\hline
Commercial System A     & 96.25          & 57.00	      & 30.83 \\
Commercial System B     & 83.67          & 63.33          &	\textbf{10.00} \\
\hline
\end{tabular}
\end{table}

\subsection{Document-Level vs. Sentence-Level Translation}\label{sub:document-vs-sentence}
From Table \ref{tab:rnd-results}, we calculate percentages indicating when the document-level score exceeds the sentence-level score. This is done by counting where \textit{Doc} is greater than the \textit{Sent} column for each row and metric.
Specifically, 60\% favor document-level in BLEU, 100\% in COMET, while DA-BERT dissents with only 15\% preferring the document-level. Thus, our automatic evaluation indicates that document-level is better than sentence-level for ChatGPT. 
This could be attributed to the fact that document-level translation preserves context more effectively when compared to sentence-level translation (See Appendix \ref{appendix:samples}).

\begin{table*}[htbp]
\centering
\caption{Average score of ChatGPT vs Commercial MT Systems}
\small 
\label{tab:rnd-engines}
\begin{tabular}{lcccccc}
\hline
\multirow{2}{*}{System} & \multicolumn{3}{c}{Document Level} & \multicolumn{3}{c}{Sentence Level}\\
&  BLEU &  COMET & DA-BERT &  BLEU &  COMET & DA-BERT\\
\hline
ChatGPT-3.5 Simple   & 25.07 & 	0.867 & \textbf{0.175} & 25.28 & 0.856 & 0.174\\
ChatGPT-4 Simple     & 25.68 & 	0.866 & 0.169 & 26.85 & 0.853 & 0.172\\
ChatGPT-3.5 Enhanced & 25.45 & 	\textbf{0.868} & 0.170 & 22.47 & 0.840 & \textbf{0.193}\\
ChatGPT-4 Enhanced   & 23.32 &  0.853 & 0.161 & 21.65 & 0.832 & 0.173\\
\hline
Commercial System A    & 26.21 & 	0.843 & 0.130 & 27.76 & \textbf{0.862} & 0.177\\
Commercial System B    & \textbf{33.01} &  	0.865 & 0.169 & \textbf{31.46} & 0.841 & 0.176\\
\hline
\end{tabular}
\end{table*}

\subsection{Simple vs. Enhanced Prompt}\label{sub:simple-vs-enhanced}
Similar to Subsection \ref{sub:document-vs-sentence}, we calculate the percentages where the enhanced prompt outperforms the simple prompt. 
However, as the results vary, we cannot make a direct conclusion. 
BLEU and COMET remain neutral at 50\%, while DA-BERT indicates that the enhanced prompt is better (60\%).
We observe that enhanced prompts can cause the translation output to deviate from the gold data. This may lead to lower scores, even for more appropriate translations, simply because it differs from the reference.

Additionally, even with the human evaluation scores we could not find conclusive evidence that enhanced prompts produce higher quality translations (See Table \ref{tab:rnd-human-results}). 
This could be due to the small-scale of our human evaluation and the bias towards gold data in the automated evaluations. 
Therefore, we believe larger scale human evaluations are required to accurately answer the question of simple vs. enhanced prompts.

\subsection{ChatGPT-3.5 vs. ChatGPT-4}\label{sub:3.5-vs-4}
Once again, we compute the percentages where ChatGPT-4 outperforms ChatGPT-3.5 for both document and sentence-level. 
In terms of automatic evaluation, we find that ChatGPT-4 is not superior to ChatGPT-3.5.
Specifically, for document-level, COMET and DA-BERT preferred ChatGPT-3.5, at 30\% and 40\% respectively, while BLEU remained neutral at 50\%. 
Similarly, for sentence-level, BLEU preferred ChatGPT-4 at 80\%, while COMET and DA-BERT preferred ChatGPT-3.5, at 40\% and 20\% respectively.

For the human evaluation (Table \ref{tab:rnd-human-results}), the overall MQM scores align with the findings in automatic evaluation: higher scores were observed for ChatGPT-4, suggesting that it is not superior to ChatGPT-3.5.
Moreover ChatGPT-3.5 is better in terms of accuracy, meaning that the translation is reflecting the source text better.
Conversely, ChatGPT-4 offers better fluency, meaning that it could generate translations that feel more native and easier to understand.
However, the decision of which one is better depends on the preferences of the end users. 
For tasks where conveying information without ambiguity is crucial, ChatGPT-3.5 might be more appropriate. 
On the other hand, for translations that include creative writing, \textit{e.g.} advertisements, where a natural flow is essential, ChatGPT-4 may be the preferred choice.

\vspace{0.2cm}
\subsection{Comparison to Commercial MT Systems}\label{sub:vs-off-the-shelf}
We averaged each metric across all datasets and organized it in Table \ref{tab:rnd-engines}. 
With document-level translation, we observe that all ChatGPT settings surpass System A's scores, while also performing on par or better than System B in COMET and DA-BERT.
On the other hand, the BLEU scores show the reverse trend, with all ChatGPT settings performing worse than both commercial systems.
At the sentence-level, we found similar trends to the ones observed at the document-level. With BLEU, ChatGPT performs worse than both System A and System B. 
Conversely, all variants of ChatGPT perform competitively against both commercial systems in COMET and DA-BERT.

Overall, ChatGPT shows impressive translation capabilities with two of the three evaluation metrics investigated, when compared to commercial translation engines. 
Additionally, this interpretation is also supported by the MQM scores (Table \ref{tab:rnd-human-results}), where three of the four investigated ChatGPT settings perform better than System A and competitively against System B. Specifically, ChatGPT-3.5 achieved better accuracy scores than System B. 
In terms of fluency, all ChatGPT settings achieved scores within the range of the well-regarded translation engines.
These are evidence that ChatGPT, with scores on par with commercial translation systems, can be used to generate translations that remain faithful to the source text.

Another important factor that we observed is that ChatGPT is slower than both commercial engines. 
We were able to mitigate this issue by using Azure's Japan East servers for ChatGPT, but it was still noticeably slower.

\section{Conclusion}

In our study comparing ChatGPT-3.5 and ChatGPT-4 for JA-EN translation, using simple and enhanced prompts, we uncovered key insights. 
Firstly, translating entire documents proved more effective than translating sentence by sentence, potentially due to enhanced contextual preservation. 
Secondly, the question of whether enhanced prompts are superior remains uncertain, as both automated and human evaluations provided inconclusive results. 
Thus, we leave it open for future researchers to answer this question through larger scale experiments.
Thirdly, automatic and human evaluations indicated an overall preference for ChatGPT-3.5, but we also point out that the end user may want to tradeoff between accuracy (ChatGPT-3.5) and fluency (ChatGPT-4).
Finally, in JA-EN translation, all investigated ChatGPT settings proved to be strong competitors against two commercial MT systems in both automatic and human assessments.

\clearpage
\bibliographystyle{unsrt}
\bibliography{main} 

\clearpage
\appendices
\section{Datasets} \label{appendix:dataset}

\begin{table}[h]
\centering
\caption{Description of Datasets}
\small
\label{tab:method-dataset}
\begin{tabular}{llcccc}
\hline
\multirow{2}{*}{Dataset} & \multirow{2}{*}{Direction} & \multirow{2}{*}{Type} & Avg.Num         & Avg.Num       \\
        &           &                & Src.Sent     & Src.Sent \\
\hline
ParaNatCom  & EN $\rightarrow$ JA &  Abstracts  & 1179.2    &   7.2  \\
FLORES      & EN $\rightarrow$ JA &  Article    & 600.6     &   5.0  \\
Novels      & EN $\rightarrow$ JA &  Novels     & 1640.6    &   16.2 \\
KFTT        & JA $\rightarrow$ EN &  History    & 742.6     &   15.0 \\
WMT News    & JA $\rightarrow$ EN &  News       & 375.0     &   7.0  \\
\hline
\end{tabular}
\end{table}

\section{Enhanced Prompts: Categories, Styles, and Tones} \label{appendix:enhanced-prompt}

For \textit{categories} and its \textit{style} mapping, the options are as follows:
\begin{enumerate}
    \item finance, economy, judicial affairs: 
        contracts, 
        prospectus, 
        financial reports, 
        research reports, 
        articles of incorporation, 
        certified copies, 
        business letters, 
        press releases.
    \item medicine: 
        application documents (CTD), 
        package inserts, 
        medical records,	
        regulatory guidelines,
        academic papers, literature.
    \item industry, science technology:
        instruction manuals,
        catalogs,
        brochures,	
        design documents,
        drawings,
        proposals,	
        regulations,	
        academic papers, literature.
	\item patent, intellectual properties:
        application documents (specifications),
        gazettes,
        prosecution,
        IDS (Information Disclosure Statement).
    \item other:
        catalogs,
        brochures,
        operations manuals,
        guidebooks,	
        article,	
        manga,
        novel.	
\end{enumerate}

For \textit{tones}, the options are as follows: \textit{spoken formal}, \textit{spoken casual}, \textit{written formal}, \textit{written casual}, \textit{plain tone (da \& de aru)}, and \textit{polite tone (desu \& masu)}.

\vspace{0.2cm}
\section{Translation Samples} \label{appendix:samples}
We sampled one source sentence from WMT News, the corresponding gold translation, and the translation outputs of all ChatGPT settings and of the two commercial translation systems for both document and sentence-level.
\vspace{0.2cm}
\\
\textbf{Source Text}: ラーメンとたいめしのおむすびを試食し「タイのだしがよく効いている。おいしい」と太鼓判を押した。
\vspace{0.2cm}
\\
\textbf{Gold}: He sampled the ramen and rice balls and said, "You can taste the sea bream soup stock and it's delicious." giving it a thumbs up.
\\
\textbf{Document-level Translation}:
\begin{itemize}
    \item \textbf{ChatGPT-3.5 Simple}: He praised the ramen and rice ball with sea bream, saying, "The Thai broth is well flavored. It's delicious."
    \item \textbf{ChatGPT-4 Simple}: After sampling the ramen and Taimeshi onigiri, he added, "The Thai broth is very effective. It's delicious," giving it a stamp of approval.
    \item \textbf{ChatGPT-3.5 Enhanced}: He praised the taste of the ramen and taimeshi rice ball after trying them, stating, "The Thai broth is well accentuated. It's delicious."
    \item \textbf{ChatGPT-4 Enhanced}:  After a taste test of the ramen and tai-meshi onigiri, he heartily approved, stating: "The snapper soup really comes through. Tastes great!" The new products will also be available at stores in neighbouring Aichi and Gifu prefectures.
    \item \textbf{System A}: Tasting the ramen and taimeshi rice balls, he said, "The Thai dashi (soup stock) is very effective. They are delicious.   
    \item \textbf{System B}: He sampled the ramen and taimeshi rice balls and praised them, saying, ``The Thai soup stock works well.It's delicious.''
\end{itemize}

\textbf{Sentence-level Translation}:
\begin{itemize}
    \item \textbf{ChatGPT-3.5 Simple}: I tasted ramen and an onigiri with sea bream and praised it, saying, "The Thai broth is very flavorful. It's delicious." 
    \item \textbf{ChatGPT-4 Simple}: They sampled the ramen and rice balls with sea bream and affirmed, "The Thai broth is very well done. It's delicious."
    \item \textbf{ChatGPT-3.5 Enhanced}: I tried tasting ramen and tai meshi onigiri and gave it my stamp of approval, saying "The Thai broth is really effective. Delicious!"
    \item \textbf{ChatGPT-4 Enhanced}: I taste-tested Ramen and Tai-meshi rice balls, and I gotta say, the Thai flavors are really coming through. It's super tasty, big thumbs up from me!
    \item \textbf{System A}: After tasting the ramen and taimeshi rice balls, the chef gave his stamp of approval, saying, "The Thai dashi (soup stock) is very effective and delicious.   
    \item \textbf{System B}: He sampled the ramen and taimeshi rice balls and praised them, saying, ``The Thai soup stock works well.It's delicious.''
\end{itemize}

\end{CJK}
\end{document}